%% file: main.tex
\documentclass[10pt,twocolumn,letterpaper]{article}

\usepackage[pagenumbers]{cvpr}
\usepackage{bm}
\usepackage{multirow}
\usepackage{colortbl}
\usepackage{comment}


\definecolor{cvprblue}{rgb}{0.21,0.49,0.74}
\usepackage[pagebackref,breaklinks,colorlinks,allcolors=cvprblue]{hyperref}

\title{Taylor Outlier Exposure}

\author{Kohei Fukuda\\
Hiroshima University, Japan\\
{\tt\small kohei.fukuda41@gmail.com}
\and
Hiroaki Aizawa\\
Hiroshima University, Japan\\
{\tt\small hiroaki-aizawa@hiroshima-u.ac.jp}
}

\begin{document}
\maketitle
\input{sec/0_abstract}    
\input{sec/1_intro}
\input{sec/2_related}
\input{sec/3_motivation}
\input{sec/4_method}
\input{sec/5_evaluation}
\input{sec/6_discussion}
\input{sec/7_conclusion}

{
    \small
    \bibliographystyle{ieeenat_fullname}
    \bibliography{main}
}

\end{document}

%% file: sec/0_abstract.tex
\begin{abstract}
Out-of-distribution (OOD) detection is the task of identifying data sampled from distributions that were not used during training. This task is essential for reliable machine learning and a better understanding of their generalization capabilities. Among OOD detection methods, Outlier Exposure (OE) significantly enhances OOD detection performance and generalization ability by exposing auxiliary OOD data to the model. However, constructing clean auxiliary OOD datasets, uncontaminated by in-distribution (ID) samples, is essential for OE; generally, a noisy OOD dataset contaminated with ID samples negatively impacts OE training dynamics and final detection performance. Furthermore, as dataset scale increases, constructing clean OOD data becomes increasingly challenging and costly. To address these challenges, we propose Taylor Outlier Exposure (TaylorOE), an OE-based approach with regularization that allows training on noisy OOD datasets contaminated with ID samples. Specifically, we represent the OE regularization term as a polynomial function via a Taylor expansion, allowing us to control the regularization strength for ID data in the auxiliary OOD dataset by adjusting the order of Taylor expansion. In our experiments on the OOD detection task with clean and noisy OOD datasets, we demonstrate that the proposed method consistently outperforms conventional methods and analyze our regularization term to show its effectiveness. Our implementation code of TaylorOE is available at \url{https://github.com/fukuchan41/TaylorOE}.
\end{abstract}

%% file: sec/1_intro.tex
\section{Introduction}
\label{sec:intro}

\begin{figure}[t]
    \centering
    \includegraphics[width=1.0\linewidth]{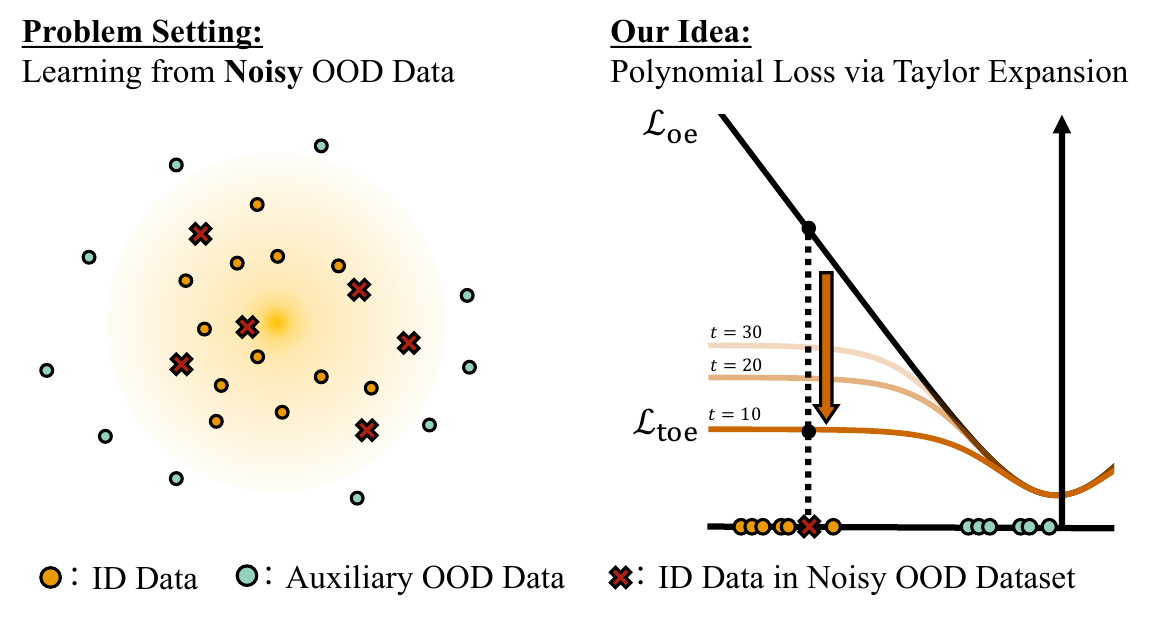}
    \caption{Overview of problem setting and our idea. We consider an auxiliary OOD dataset for OE that includes not only OOD data~(blue circles) but also ID data~(red crosses). Our TaylorOE $\mathcal{L}_{\text{toe}}$, a polynomial loss function derived from Taylor expansion, suppresses the influence of the ID data in the noisy OOD dataset by adjusting the regularization strength based on the order $t$.}
    \label{fig:overview}
\end{figure}

Out-of-distribution (OOD) detection is the task of identifying data sampled from distributions that were not used during training. This capability is fundamental for achieving model generalization and establishing reliable machine learning models. To accomplish this task, numerous sophisticated methods have been proposed, including uncertainty-based approaches~\cite{baseline,mahalanobis,odin,maxlogit,gradnorm,dice,ash}, confidence-aware techniques~(\eg, confidence prediction~\cite{conf_learning} and overconfidence mitigation~\cite{logitnorm}), and approaches leveraging the complexity of OOD data~\cite{mood}.

Among the various methods, the most effective is outlier exposure (OE), proposed by Hendrycks~\etal~\cite{outlier_exposure}. The idea behind OE is to expose auxiliary OOD data to the model, thereby enhancing the model’s OOD detection capabilities. Specifically, OE adds a regularization term that aligns the predicted softmax probabilities for the OOD data with a uniform distribution, in addition to the cross-entropy loss calculated from the in-distribution (ID) data. Inspired by this simple yet effective idea, various follow-up methods have been proposed, including enhancements to the regularization of auxiliary OOD data~\cite{energy}, sampling strategies for auxiliary OOD data~\cite{resampling,atom,poem,dos,grad_reg}, and augmentation methods for small amounts of auxiliary OOD data~\cite{divoe,doe,MixOE}. 

However, OE-based methods require a sufficiently diverse auxiliary OOD dataset to cover the OOD space and a clean dataset constructed from OOD data uncontaminated by ID samples. Generally, OE training on a noisy or low-quality auxiliary OOD dataset adversely affects training dynamics and degrades OOD detection performance~(see Sec.~\ref{analysis}). Moreover, as dataset scale increases, building high-quality and clean auxiliary OOD datasets becomes challenging and costly, posing a critical challenge for OE and follow-up methods.

\begin{figure}[t]
    \centering
    \includegraphics[width=1.0\linewidth]{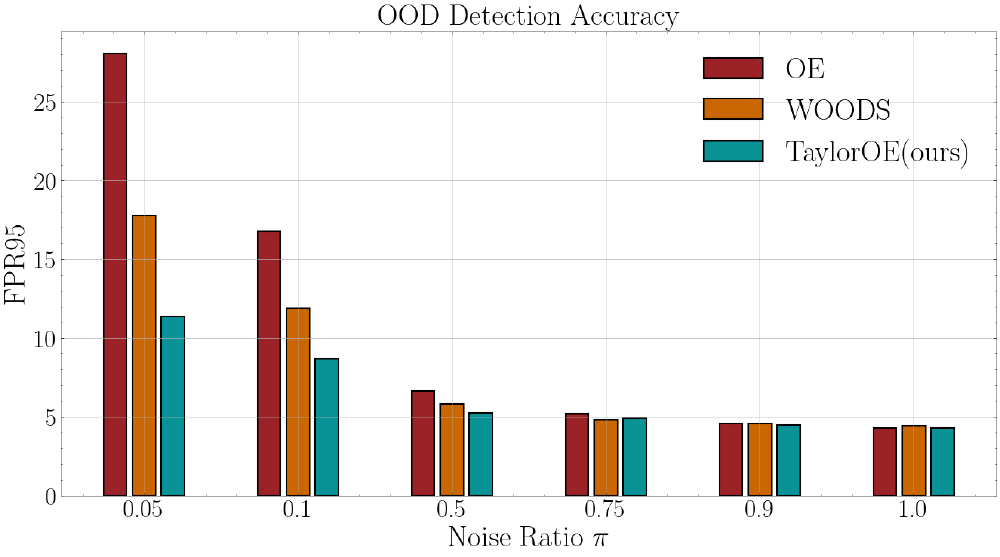}
    \caption{Comparison of TaylorOE with OE~\cite{outlier_exposure} and WOODS~\cite{woods}. The noise ratio $\pi$ represents the proportion of clean OOD data in the auxiliary OOD dataset. TaylorOE outperforms these conventional methods across various noise ratios, demonstrating its ability to work effectively even with noisy data.}
    \label{fig:overview_results}
\end{figure}

Therefore, we aim to enable OE to work effectively even when using a noisy auxiliary OOD dataset, as shown on the left of Fig.~\ref{fig:overview}. From our preliminary experiments investigating the values of OE regularization term for ID and OOD data, we found that ID data have relatively low values, while many clean OOD samples exhibit comparatively high values~(see Sec.~\ref{analysis}). Based on this analysis, we propose Taylor Outlier Exposure (TaylorOE), which incorporates a polynomial expression of the OE regularization term. TaylorOE is represented by a finite-dimensional Taylor expansion of the OE term as the regularization term for OOD data. TaylorOE enables control over regularization strength by adjusting the order of the Taylor expansion, thereby maintaining the influence of OOD data while suppressing the impact of ID data in the OOD dataset, as shown in the right of Fig.~\ref{fig:overview}. In our experiments, we compared conventional methods with TaylorOE on noisy auxiliary OOD datasets with various noise ratios. As shown in Fig.~\ref{fig:overview_results}, we demonstrate that the proposed method works effectively even on noisy auxiliary OOD datasets contaminated by ID data and consistently outperforms conventional methods.

Our main contributions are as follows:
\begin{itemize}
    \item We introduce a polynomial regularization of outlier exposure via Taylor expansion, enabling adaptive control of the regularization strength based on the ID and OOD data characteristics.
    \item We demonstrate that the proposed method outperforms conventional methods, particularly in cases with a higher noise ratio. TaylorOE can learn OOD detection ability from noisy auxiliary OOD data without requiring data cleaning or additional OOD sampling.
    \item We show that TaylorOE is easily applicable to other OE-based techniques, such as OOD synthesis~\cite{divoe} and OOD resampling~\cite{resampling}.
\end{itemize}

%% file: sec/2_related.tex
\section{Related Work}
\label{sec:related}

\subsection{Out-of-Distribution Detection}
\label{sec:ood}

OOD detection is critically important for ensuring the reliability and safety of machine learning models. Various approaches have been proposed for improving OOD detection accuracy, such as refining OOD score functions~\cite{baseline,mahalanobis,odin,maxlogit,gradnorm,dice,ash}, controlling confidence~\cite{conf_learning,logitnorm}, considering the complexity of OOD data~\cite{mood} and leveraging auxiliary OOD data~\cite{outlier_exposure}. Among such approaches, OE~\cite{outlier_exposure}, a promising approach, employs a real OOD dataset to enhance OOD detection performance by maximizing the entropy of the auxiliary OOD data. However, the use of real OOD data remains a critical challenge related to computational cost, the collection of OOD data, and annotation cost.

\noindent\textbf{Computational cost:} The auxiliary OOD data must cover the OOD space comprehensively, resulting in a large dataset size. To efficiently train the model on such a dataset, a sampling strategy~\cite{resampling,atom,poem,dos,grad_reg} selects critical OOD data in the auxiliary OOD dataset to improve the decision boundary between ID and OOD data.

\noindent\textbf{Data collection cost:} OOD data are typically sampled from unknown distributions, and due to their diversity, constructing large and high-quality datasets is challenging. Consequently, it is essential to enhance detection accuracy using only a limited amount of auxiliary OOD data. To address this issue, methods have been proposed for augmenting a small amount of auxiliary OOD data by properly extrapolating input data~\cite{divoe} or by applying transformations through perturbations to the model\cite{doe}.

\noindent\textbf{Annotation cost:} Most OOD detectors, including OE-based methods and those that use only ID data for OOD detection, require distinguishing ID and OOD data beforehand. However, due to the large scale of the dataset and the characteristics of OOD data, annotation is often time-consuming. Some approaches have been proposed to address this issue, such as those that sample synthetic auxiliary OOD data in feature space~\cite{vos,npos}, generate OOD images~\cite{dream}, and utilize open world data~\cite{woods,isun}.

In this paper, with regard to data collection and annotation costs, we start from the assumption that constructing a clean auxiliary OOD dataset is impractical. Therefore, we aim to train OE using noisy auxiliary OOD datasets contaminated by ID data. Our assumption for learning with noisy OOD datasets is close to the work of Katz-Samuels ~\etal~\cite{woods}. Therefore, our experiments follow their setting~\cite{woods}. The proposed TaylorOE is based on the OE framework, with the key idea of relaxing regularization for the ID data in the OOD data via Taylor expansion. In addition, the proposed method can be easily applied to techniques such as OOD sample synthesis~\cite{divoe}, which reduces data collection costs, and a sampling strategy~\cite{resampling}, which reduces computational costs.

\subsection{Robust Polynomial Loss Function}
\label{sec:polyloss}

Cross-entropy loss is widely used in various classification tasks, including OOD detection; however, when the training data include noisy labels, cross-entropy loss tends to overfit noisy samples, leading to a decrease in the accuracy of the test data. To address this issue, various loss functions have been proposed~\cite{generalized_ce,peer_loss,taylor_ce}. Feng~\etal~\cite{taylor_ce} introduced a robust loss function derived from a Taylor expansion of cross-entropy loss. By truncating higher-order terms, the approximation becomes similar to the robust mean absolute error, effectively reducing the influence of noisy samples during training. This concept was further generalized in the study of Leng~\etal~\cite{polyloss}, who proposed an adaptive loss function that adjusts the polynomial coefficients to suit the dataset. In this paper, we propose a robust loss function for ID data in noisy auxiliary OOD datasets based on the Taylor expansion concept in the context of OOD detection.

%% file: sec/3_motivation.tex
\section{Problem Definition and Motivation}

\subsection{Problem Definition}
\noindent\textbf{OOD Detection}\label{setup_ood}: In this paper, we consider a $K$-class classification problem. Let $\mathcal{X} \subset \mathbb{R}^d$ denote the $d$-dimensional input space, and $\mathcal{Y} = {1, \ldots, K}$ denote the label space. We define $\mathcal{D}_{\text{in}}$ as a training dataset consisting of independent and identically input-label pairs obtained from the joint distribution $\mathcal{P}_{\mathcal{X} \times \mathcal{Y}}$, which we refer to as ID. On the other hand, a dataset obtained from classes different from the ID label space (\ie $y \notin \mathcal{Y}$) is referred to as an OOD dataset and denoted as $\mathcal{D}_{\text{out}}$. The objective of OOD detection is to calculate an OOD score $S(x)$ for each input $x \sim \mathcal{X}$ sampled from the input space based on the $K$-class prediction obtained from a model trained on either $\mathcal{D}_{\text{in}}$ alone or on both $\mathcal{D}_{\text{in}}$ and $\mathcal{D}_{\text{out}}$. This score is then used to classify the input into the ID or OOD class with a threshold $\tau$.

\noindent\textbf{Auxiliary OOD Dataset}:
As previously mentioned, the use of auxiliary OOD data is effective for OOD detection. Let this auxiliary OOD dataset be denoted as $\mathcal{D}_{\text{out}}^{\text{aux}}$. It is necessary to perform data cleaning to ensure that it contains only OOD data. This requires that $\mathcal{D}_{\text{out}}^{\text{aux}} \cap \mathcal{D}_{\text{in}} = \emptyset$, meaning that the auxiliary OOD dataset represents a set of data sampled from the OOD. In this paper, we also refer to this dataset $\mathcal{D}_{\text{out}}^{\text{aux}}$ as a clean OOD dataset.

\noindent\textbf{Noisy Auxiliary OOD Dataset}:
There are various challenges associated with using auxiliary OOD data. One challenge is the removal of ID data from noisy auxiliary OOD datasets. To reduce this preprocessing cost of auxiliary OOD data described in Sec.~\ref{sec:ood}, we aim to utilize noisy auxiliary OOD data $\mathcal{D}_{\text{out}}^{\text{noisy}}$, composed of clean OOD data and ID data, for training. Following Katz-Samuels~\etal\cite{woods} and using the Huber contamination model \cite{huber}, we define $\mathcal{D}_{\text{out}}^{\text{noisy}}$ as follows:
\begin{align}\mathcal{D}_{\text{out}}^{\text{noisy}} := (1-\pi)\mathcal{D}_{\text{in}} + \pi \mathcal{D}_{\text{out}}^{\text{aux}},
\end{align}
where $\pi \in (0,1]$. Here, $\pi$ represents the proportion of preprocessed clean OOD data in the auxiliary OOD dataset, while the proportion of ID data is $1-\pi$. In this paper, we also refer to this dataset $\mathcal{D}_{\text{out}}^{\text{noisy}}$ as a noisy auxiliary OOD dataset or a noisy OOD dataset.

\subsection{Motivation}\label{motivation}
The goal of this paper is to perform OE using a noisy auxiliary OOD dataset $\mathcal{D}_{\text{out}}^{\text{noisy}}$. Specifically, we aim to reduce the influence of ID data in the OOD dataset while maintaining the influence of the clean auxiliary OOD data. In the end, we introduce a polynomial loss function derived from the Taylor expansion of OE regularization, enabling OE training on a noisy OOD dataset.

%% file: sec/4_method.tex
\section{Methodology}
\label{sec:methodology}

\subsection{Analysis of Outlier Exposure}\label{analysis}
In this section, we explain OE~\cite{outlier_exposure} and analyze the softmax probabilities and OE's regularization term $\mathcal{L}_{\text{oe}}$ of samples obtained from $\mathcal{D}_{\text{in}}$ and $\mathcal{D}_{\text{out}}^{\text{aux}}$, which are the components of $\mathcal{D}_{\text{out}}^{\text{noisy}}$. For the analysis, we use a pretrained Wide Residual Network~\cite{wideresnet} and evaluate 10,000 samples each from the ID data, namely \texttt{CIFAR-10}~\cite{cifar} and the auxiliary OOD data, namely \texttt{300K Random Images}~\cite{outlier_exposure}.

\noindent\textbf{Preliminaries:} OE is a method that applies regularization to make the softmax probabilities approach uniformity for auxiliary OOD data that does not contain ID data. The loss function is formulated as follows:
\begin{align}\label{oe_loss}
    \mathcal{L} &= \mathbb{E}_{\{x, y\} \sim \mathcal{D}_{\text{in}}}[\mathcal{L}_{\text{ce}}(f(x),y)]\nonumber \\
    &\qquad+ \lambda \mathbb{E}_{x\sim \mathcal{D}_{\text{out}}^{\text{aux}}} [\mathcal{L}_{\text{oe}}(f(x))].
\end{align}
$f(\cdot)$ is a deep neural network trained on ID data. The first term of the loss function represents the cross-entropy loss $\mathcal{L}_{\text{ce}}$ for ID data, and the second term is the Kullback-Leibler divergence between the softmax probabilities and a uniform distribution for the auxiliary OOD data. $\lambda$ is a hyperparameter that balances these two terms. The second term $\mathcal{L}_{\text{oe}}$ is calculated using the following equation:
\begin{align}
    \label{l_oe}
    \mathcal{L}_{\text{oe}}(f(x)) = \frac{1}{K} \sum^K_{i=1}-\log p_i,
\end{align}
where $p_i$ denotes the $i$-th softmax probability.

\begin{figure*}[t]
    \centering
    \begin{minipage}{0.33\hsize}
        \centering
        \includegraphics[width=1.0\linewidth]{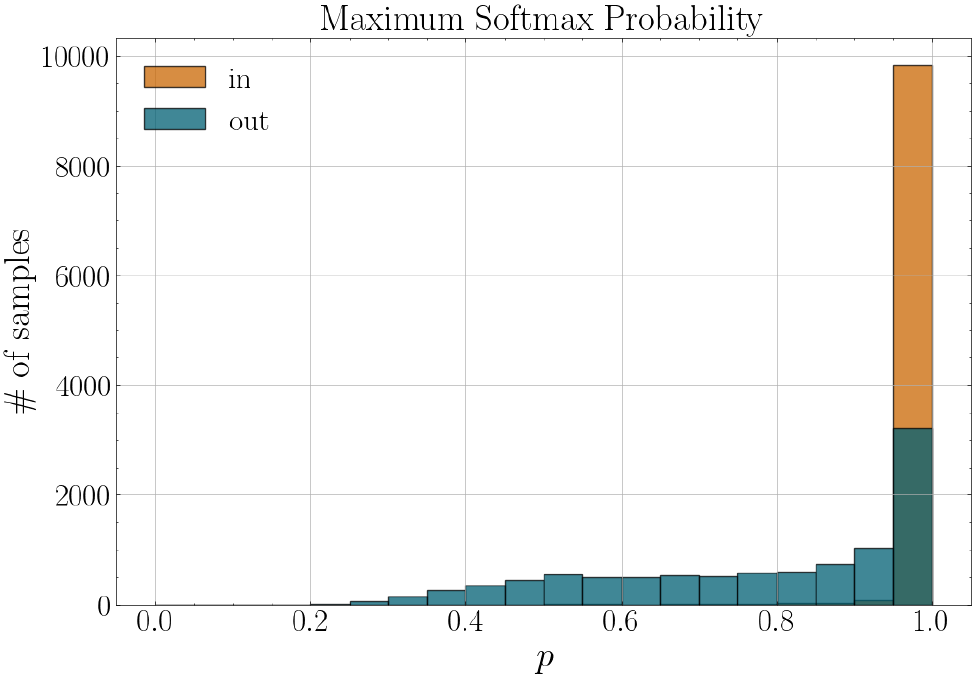}
    \end{minipage}
    \begin{minipage}{0.33\hsize}
        \centering
        \includegraphics[width=1.0\linewidth]{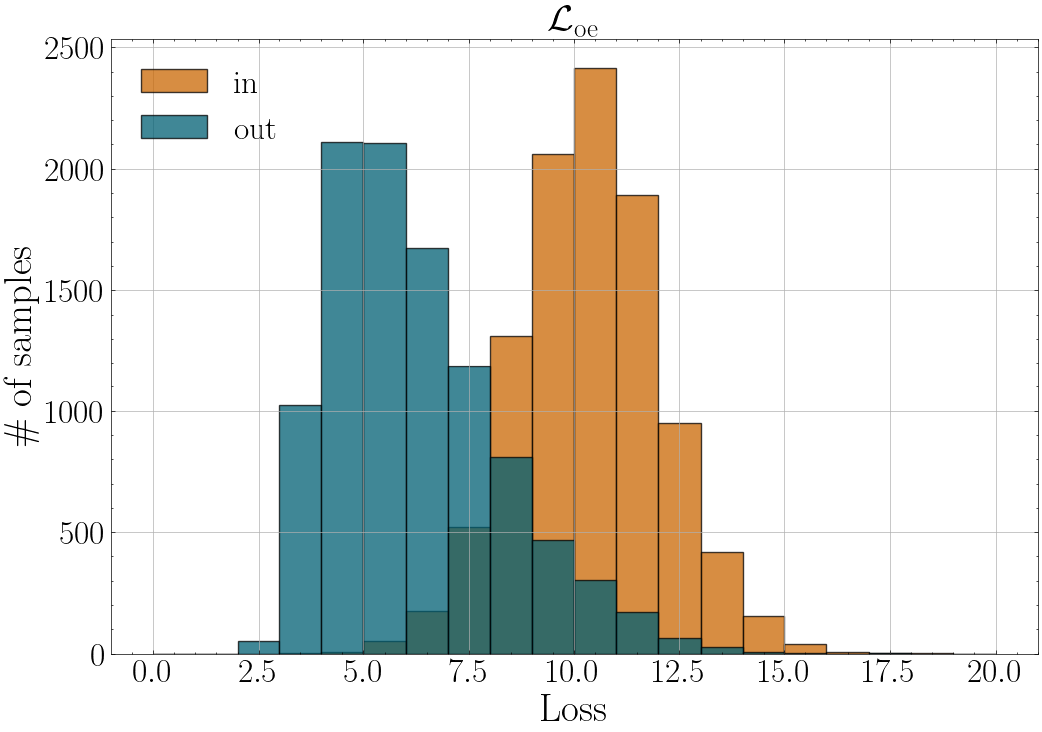}
    \end{minipage}
    \caption{Histograms of maximum softmax probabilities for ID and OOD data~(left) and Eq.~\ref{l_oe}~(right) obtained from the pretrained model on ID data. For ID data, the maximum softmax probabilities of most samples are close to 1. As a result, the probabilities of other classes are close to 0, causing the entropy to rise and leading to larger values for Eq.~\ref{l_oe}. On the other hand, clean OOD data has fewer samples with maximum softmax probabilities close to 1 compared to ID data, and probabilities are distributed across multiple classes to some extent, causing the entropy to fall and leading to smaller values for Eq.~\ref{l_oe}.}
    \label{fig:msp_entropy}
\end{figure*}

\begin{figure*}[t]
    \centering
    \begin{minipage}{0.33\hsize}
        \centering
        \includegraphics[width=1.0\linewidth]{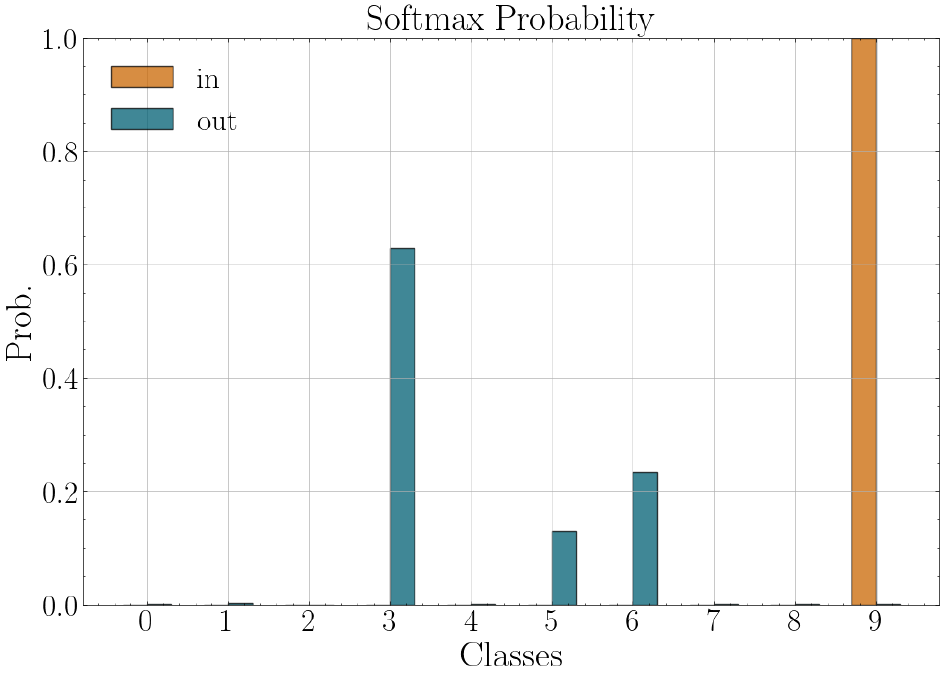}
    \end{minipage}
    \begin{minipage}{0.33\hsize}
        \centering
        \includegraphics[width=1.0\linewidth]{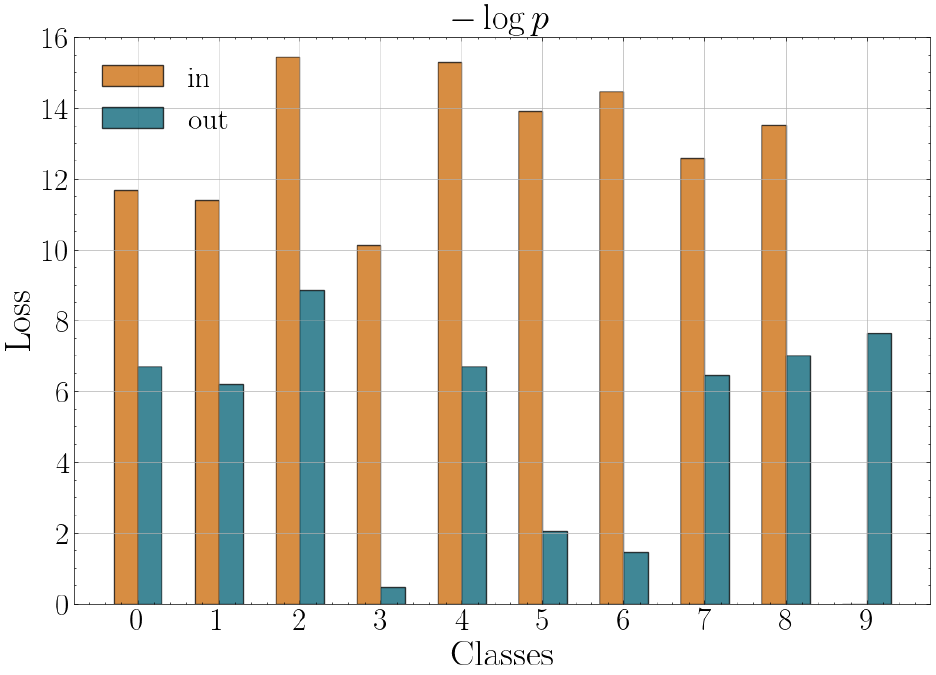}
    \end{minipage}
    \begin{minipage}{0.33\hsize}
        \centering
        \includegraphics[width=1.0\linewidth]{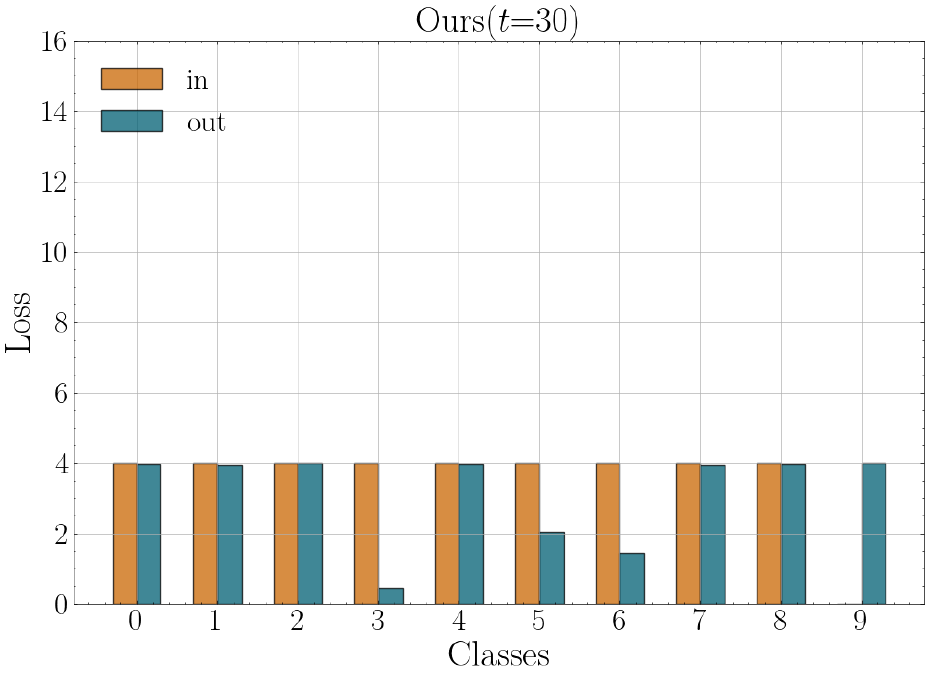}
    \end{minipage}
    \caption{Softmax probabilities for ID and OOD samples (left), $-\log p$ for each class (center), and Taylor series expansion of $-\log p$ used in the proposed method (right). For ID samples, the predicted probability is concentrated on a single class, resulting in larger values for $-\log p$ of other classes. By performing a Taylor expansion up to the finite order, the proposed method prevents the information from classes with small predicted probabilities from becoming excessively large. For OOD samples, the information from classes such as the fifth and sixth classes, which do not have excessively large information, is preserved.}
    \label{fig:loss_bar}
\end{figure*}

\noindent\textbf{Propeties of OE:}
Models trained on ID data exhibit a characteristic pattern where the maximum softmax probabilities are high for ID data but comparatively low for OOD data~\cite{baseline}. This phenomenon is demonstrated in the left plot of Fig.~\ref{fig:msp_entropy}.  We examine the softmax probabilities for each class and the magnitude of the $-\log p$ values that appear in Eq.~\ref{l_oe} by taking one sample from both the ID data and OOD data (Fig.\ref{fig:loss_bar}). For ID data, one class is assigned a probability of close to 1, while other classes are assigned values close to 0. Consequently, the $-\log p$ values, except for the class with the highest probability, become large, which results in an overall larger value for the mean in Eq.\ref{l_oe}. On the other hand, OOD data tends to have reasonably large probabilities assigned to multiple classes, leading to a smaller value for Eq.\ref{l_oe} compared to that for ID samples. As shown in the right plot of Fig.\ref{fig:msp_entropy}, the distribution of Eq.~\ref{l_oe} tends to be larger for ID data than for OOD data, indicating that the information content of ID data is higher, while that of OOD data is smaller.

Based on this analysis, to achieve our objective (Sec.~\ref{motivation}), we need to employ a regularization function in which the regularization term for samples from $\mathcal{D}_{\text{out}}^{\text{noisy}}$ with a very large maximum softmax value is reduced. In contrast, the regularization term for other samples remains close to the value of Eq.~\ref{l_oe}.

\subsection{Taylor Outlier Exposure}
In this section, we propose a regularization method for mitigating the negative effect of ID data in $\mathcal{D}_{\text{out}}^{\text{noisy}}$. We refer to this training method as TaylorOE.

In Sec.~\ref{analysis}, we found that for the ID data in $\mathcal{D}_{\text{out}}^{\text{noisy}}$, the presence of many classes with predicted probabilities of close to zero results in large $-\log p$ values, thereby increasing $\mathcal{L}_{\text{oe}}$. Therefore, we utilize the Taylor expansion to better understand the behavior of $-\log p$. The Taylor expansion represents a function as a polynomial, with coefficients calculated from its derivatives. The result of the Taylor expansion of $-\log p$ at $p=1$ is as follow:
\begin{align}\label{taylor_logp}
    -\log p=\sum^{\infty}_{n=1} \frac{(1-p)^n}{n!}.
\end{align}

\begin{figure}[t]
    \centering
    \includegraphics[width=0.85\linewidth]{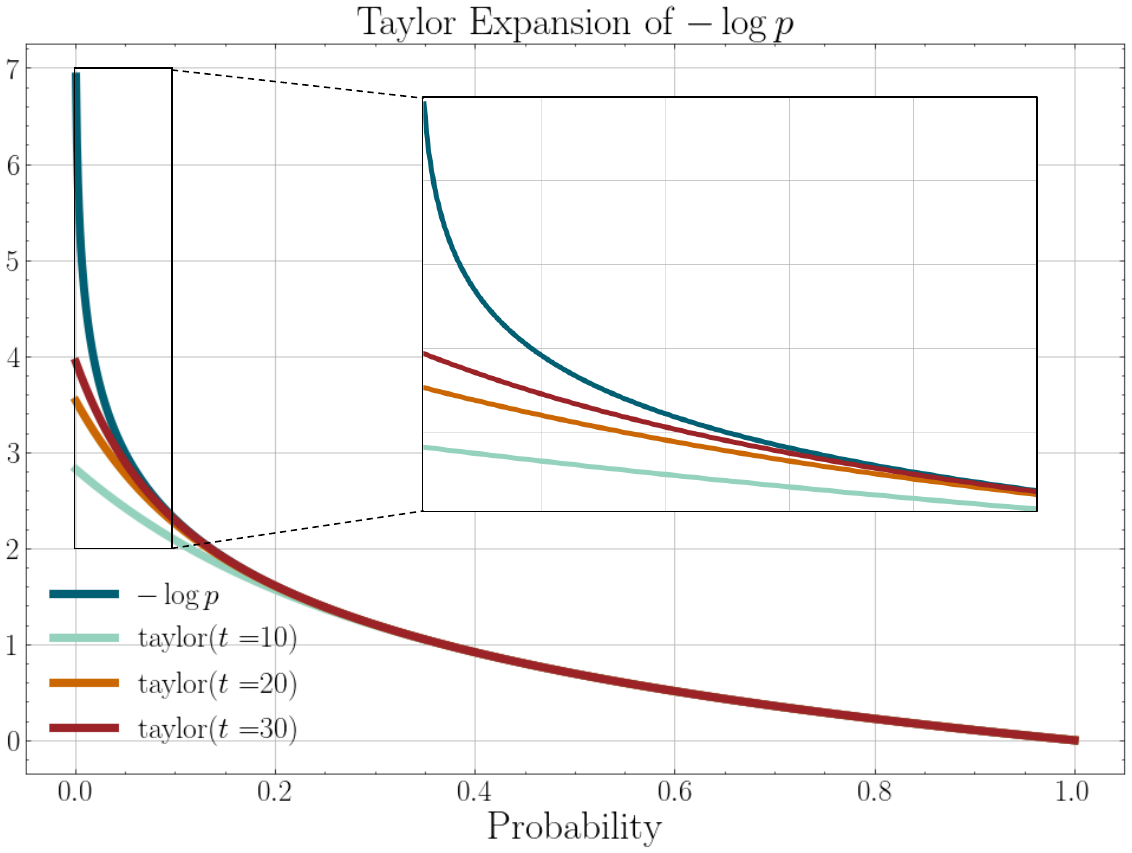}
    \caption{Taylor expansion of $-\log p$ around $p=1$. The polynomial function around $p = 1$ closely matches the original function $-\log p$ near $p = 1$ and lies below the original function near $p = 0$. Additionally, as the finite order of the expansion increases, the value near $p = 0$ gradually approaches the original function.}
    \label{fig:taylor_log}
\end{figure}

By setting $n \rightarrow \infty$ to finite order $t$, this function becomes almost equal to $-\log p$ around $p=1$ and smaller than $-\log p$ around $p=0$, as shown in Fig~\ref{fig:taylor_log}. The property of this function is shown in Fig.~\ref{taylor_logp}. TaylorOE replaces the regularization term $\mathcal{L}_{\text{oe}}$ of OE with a new regularization term $\mathcal{L}_{\text{toe}}$, based on Eq.~\ref{taylor_logp}, which is expressed as follows:
\begin{align}
    \mathcal{L}_{\text{toe}} = \frac{1}{K} \sum^{K}_{i=1} \sum^{t}_{n=1} \frac{(1-p_i)^n}{n!}.
    \label{l_toe}
\end{align}
The right plot of Fig.~\ref{fig:loss_bar} shows $\mathcal{L}_{\text{toe}}$ computed by the proposed method. The larger part in $-\log p$ is smaller so that the training is not negatively affected by the ID samples in the OOD dataset.

\begin{figure}[t]
    \centering
    \begin{minipage}{0.49\hsize}
        \centering
        \includegraphics[width=1.0\linewidth]{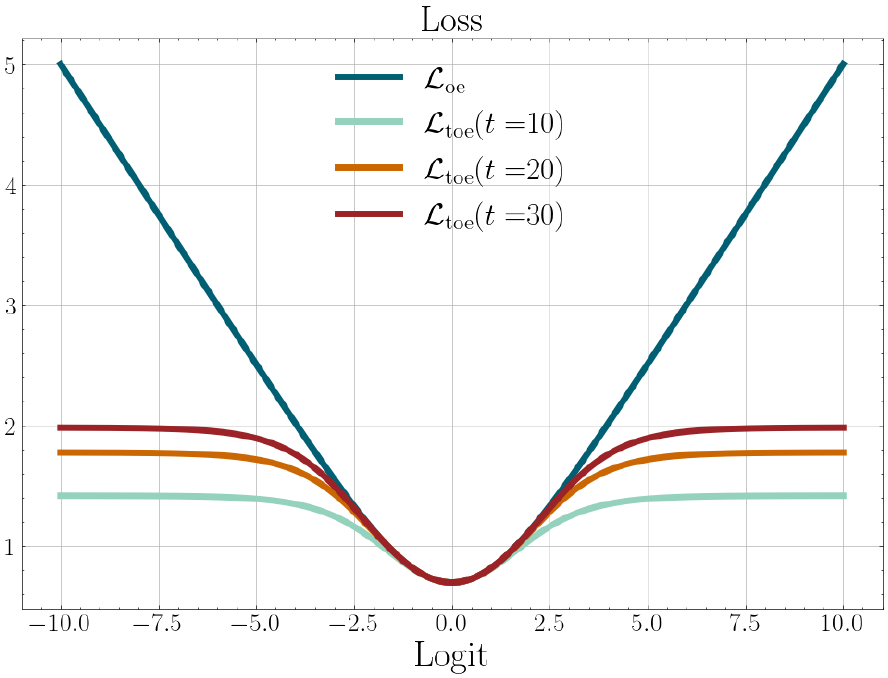}
    \end{minipage}
    \begin{minipage}{0.49\hsize}
        \centering
        \includegraphics[width=1.0\linewidth]{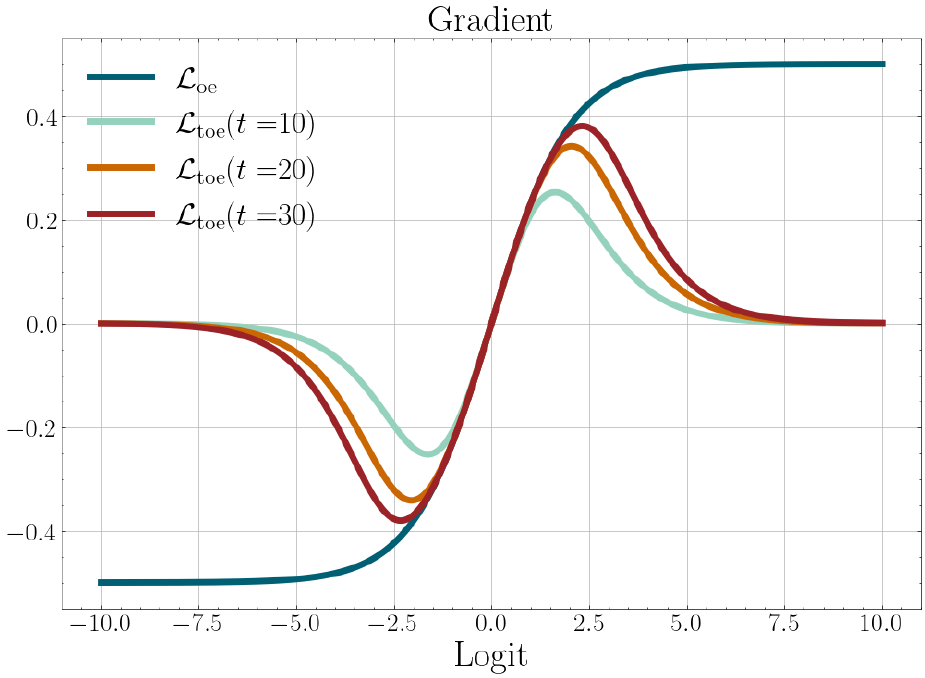}
    \end{minipage}
    \caption{Relationship between logits and $\mathcal{L}_{\text{oe}}$~(Eq.~\ref{l_oe}) and $\mathcal{L}_{\text{toe}}$~(Eq.~\ref{l_toe}). As the logit increases, $\mathcal{L}_\text{oe}$ and the gradient magnitude both become larger, thereby having a greater impact on OE training. In contrast, $\mathcal{L}_{\text{toe}}$ does not increase excessively for very large logits and the gradient becomes zero. This demonstrates that $\mathcal{L}_{\text{toe}}$ can reduce the impact of learning from samples with excessively large logits, such as ID data in the auxiliary OOD dataset.}
    \label{fig:logit_graph}
\end{figure}

\noindent\textbf{Intuitive Understanding of TaylorOE:}
To provide a more intuitive understanding of the proposed method, we consider binary classification based on a single logit output, similar to logistic regression. In this case, if one logit is $z$, the logit of the other class is $-z$. Fig.~\ref{fig:logit_graph} shows the values of $\mathcal{L}_{\text{oe}}$ and $\mathcal{L}_{\text{toe}}$ (left) and their gradients (right) with respect to the logits. For $\mathcal{L}_{\text{oe}}$, the loss is minimized when the logit is zero; it increases as the logit deviates from zero. The same observation holds for the gradient magnitude: as the difference between class logits $z-(-z)$ increases, resulting in the impact on training. On the other hand, for $\mathcal{L}_{\text{toe}}$, the loss behaves similarly to that for $\mathcal{L}_{\text{oe}}$ when the logit is near zero. However, beyond a certain logit value from zero, the loss value becomes almost constant, and the gradient approaches zero. This indicates that the influence of training becomes negligible when the difference between class logits becomes very large. Increasing the finite order $t$ of $\mathcal{L}_{\text{toe}}$ relaxes the threshold at which training influence becomes negligible, with this regularization function $\mathcal{L}_{\text{toe}}$ gradually converging to $\mathcal{L}_{\text{oe}}$.

%% file: sec/5_evaluation.tex
\section{Evaluation}
\label{sec:evaluation}

\subsection{Evaluation Settings}

\noindent\textbf{Datasets:} Following the benchmark of OOD detection in image classification, we use \texttt{CIFAR-10} and \texttt{CIFAR-100}~\cite{cifar} for ID data. We also use \texttt{300K Random Images}, a subset of \texttt{80 Million Tiny Images}, as auxiliary OOD data. In addition, to measure the OOD detection accuracy after training, \texttt{SVHN}~\cite{svhn}, \texttt{LSUN-R}, \texttt{LSUN}~\cite{lsun}, \texttt{iSUN}~\cite{isun}, \texttt{Texture}~\cite{texture}, and \texttt{places365}~\cite{places365} are used as test OOD data. From the ID data, 25,000 images are used as ID data in the noisy auxiliary OOD data and the remaining 25,000 images are used as regular ID data. For each iteration, we create mini-batches from the noisy auxiliary OOD dataset, with the ratio of ID data to clean auxiliary OOD data set to $(1-\pi): \pi$. The construction of the contaminated auxiliary OOD data thus follows the approach outlined in \cite{woods}.

\noindent\textbf{Evaluation Metrics:} To measure OOD detection performance, we use FPR95 and the area under the receiver operating characteristic curve(AUROC). These metrics are commonly used to measure OOD detection accuracy.

\input{tables/main_results_cifar10}
\input{tables/main_results_cifar100}

\noindent\textbf{Training Procedure:} The model used in the experiments was a Wide Residual Network~(WRN-40-2)\cite{wideresnet}, which is the standard model used in various studies~\cite{outlier_exposure,energy}. We initialized the parameters using a model pretrained on ID data and then updated the network parameters using Nesterov's accelerated gradient descent method with a momentum of $0.9$ and a weight decay of $0.0005$, following the previous research work~\cite{outlier_exposure,energy}. Both \texttt{CIFAR-10} and \texttt{CIFAR-100} were trained for 50 epochs, with the learning rate initially set to $0.001$ and then decayed using a cosine learning rate scheduler~\cite{cosine_scheduler}. The batch size was set to 128 for both ID and OOD data. The hyperparameters for each method were determined by splitting the training data into training and validation sets and tuning based on FPR95 values on the validation data. For the proposed method, the finite order $t$ of the Taylor expansion was selected from $\{30, 60, 90, 120, 150, 180, 210, 240, 270, 300\}$. The experimental results presented below are averages based on five different random seeds. For comparison, the OOD scoring function used throughout is the energy score~\cite{energy}.

\subsection{Main Results}
In this section, we compare OE~\cite{outlier_exposure} with the proposed method, TaylorOE, to evaluate how effectively contaminated auxiliary OOD data can be utilized in training with a simple regularization-based approach. The comparison results for \texttt{CIFAR-10} as ID data are shown in Table~\ref{tb:main_results_cifar10} and those for \texttt{CIFAR-100} are shown in Table~\ref{tb:main_results_cifar100}. The ratio $\pi$, representing the proportion of ID data to clean OOD data in the auxiliary OOD dataset, was set to $\{0.05, 0.1, 0.5, 0.75, 0.9, 1.0\}$. In the tables, values given in bold indicate better performance.

\noindent\textbf{OE with Noisy OOD Dataset:}
First, we examine how OE's results vary with changes in $\pi$. As shown in Table~\ref{tb:main_results_cifar10} and Table~\ref{tb:main_results_cifar100}, when $\pi$ is small, indicating a high proportion of ID data in the auxiliary OOD dataset, OE's OOD detection performance is poor. However, as $\pi$ is larger, the performance improves, as this condition approaches the assumptions of OE. Thus, it can be understood that the ID samples in the auxiliary OOD dataset exert a negative impact on OE training, as they have larger gradients compared to those of clean OOD data and thus more strongly align with a uniform distribution. This phenomenon indicates that ID samples contained in the auxiliary OOD dataset have a negative impact on OE training. To address this issue, TaylorOE relaxes regularization to mitigate this negative impact.

\noindent\textbf{TaylorOE with Noisy OOD Dataset:}
 As shown in Table~\ref{tb:main_results_cifar10}, when $\pi$ is small, the proposed method significantly outperforms OE in detection accuracy. Specifically, substantial improvements are observed on datasets \texttt{iSUN} and \texttt{LSUN-R}, with a maximum FPR95 improvement of 23.73\% at $\pi=0.05$. The average improvement is 16.68\%. These results demonstrate that the proposed method effectively mitigates the negative impact of ID data in the auxiliary OOD dataset on the OE training process. This is because the ID data, which have low information content, tend to align with a uniform distribution strongly, reducing the degree of separation between ID and OOD data compared to when the data are clean. On the other hand, as $\pi$ is large, the performance difference between the two methods narrows. This occurs because the proposed method approximates OE under these conditions. The same trend was observed with \texttt{CIFAR-100}.

\input{tables/various_results2}
\subsection{Performance Comparison}
In this section, we compare energy obtained using a method without fine-tuning~\cite{energy}, OE~\cite{outlier_exposure}, WOODS~\cite{woods}, and TaylorOE with the present experimental setup to examine whether fine-tuning is necessary when using TaylorOE with any auxiliary OOD dataset. Table~\ref{tb:various_results} shows the average detection accuracy of six OOD datasets for various values of $\pi$. OE achieves very high accuracy when $\pi$ is large, but it is less effective when $\pi$ is small. WOODS and TaylorOE outperform OE and the method without fine-tuning, showing enhanced detection accuracy even with small $\pi$ values. These results demonstrate the benefit of using $\mathcal{D}_{\text{out}}^{\text{noisy}}$. Comparing WOODS and TaylorOE, the proposed method generally achieves better detection accuracy, with significant improvements, especially at $\pi = 0.05$ and $0.1$ (up to 9.02\% improvement in FPR95).

\input{tables/extended_oe_results2}
\subsection{Application to Advanced OE Methods}\label{sec:extendedOE}
We discussed the effectiveness of TaylorOE under various noise conditions. The TaylorOE method, which utilizes Taylor expansion to relax the negative impact of ID data in the auxiliary OOD dataset, can be easily applied to various OE extensions by replacing $\mathcal{L}_{\text{oe}}$ with $\mathcal{L}_{\text{toe}}$. We applied TaylorOE to advanced OE methods: Resampling~\cite{resampling} which efficiently resamples auxiliary OOD data, and DivOE~\cite{divoe}, which synthesizes auxiliary OOD data to expose a broader range of OOD samples. In the experiment, we used 25,000 ID samples for training and constructed an auxiliary OOD dataset of 50,000 samples consisting of the remaining 25,000 ID samples and 25,000 randomly selected samples from \texttt{300K Random Images}. In this setting, the ID-to-OOD ratio in the auxiliary OOD dataset was $0.5:0.5$ (\ie $\pi=0.5$). Table~\ref{tb:extended_oe_results} shows the results obtained with Resampling~\cite{resampling} and DivOE~\cite{divoe}.

\noindent\textbf{OOD resampling:}
We now discuss the effectiveness of applying TaylorOE to OOD resampling methods for auxiliary OOD data training. In Table~\ref{tb:extended_oe_results}, standard OE often sees no improvement or even a decrease in accuracy with Resampling~\cite{resampling}, whereas TaylorOE has better performance with Resampling compared to the Normal setting. This indicates that TaylorOE can leverage the benefits of Resampling even with the noisy auxiliary OOD dataset.

\noindent\textbf{OOD synthesis:}
We now discuss the effectiveness of applying TaylorOE to OOD synthesis methods that improve the decision boundary between ID and OOD data. As shown in Table~\ref{tb:extended_oe_results}, DivOE with TaylorOE successfully prevented the deterioration of OOD detection accuracy observed with OE, but the performance did not consistently surpass that obtained with the Normal setting. We believe that this outcome is due to TaylorOE diminishing the influence of hard samples on learning, which may not align well with DivOE's approach of generating challenging samples.

%% file: tables/main_results_cifar10.tex
\begin{table*}[t]
\centering
\caption{Comparison between OE and TaylorOE with \texttt{CIFAR-10} as ID data.}
\label{tb:main_results_cifar10}
\scalebox{0.7}{
\begin{tabular}{l|cccccccccccc}
\hline
\multicolumn{1}{c|}{\multirow{3}{*}{\textbf{Method}}} & \multicolumn{12}{c}{\textbf{OOD Dataset}}                                                                                                                                                                                                                                                                                                                                                                                                                             \\
\multicolumn{1}{c|}{}                                 & \multicolumn{2}{c}{\textbf{SVHN}}                                         & \multicolumn{2}{c}{\textbf{LSUN-R}}                                       & \multicolumn{2}{c}{\textbf{LSUN-C}}                                       & \multicolumn{2}{c}{\textbf{iSUN}}                                         & \multicolumn{2}{c}{\textbf{Texture}}                                      & \multicolumn{2}{c}{\textbf{Places365}}                                    \\
\multicolumn{1}{c|}{}                                 & \multicolumn{1}{l}{FPR95$\downarrow$} & \multicolumn{1}{l}{AUROC$\uparrow$} & \multicolumn{1}{l}{FPR95$\downarrow$} & \multicolumn{1}{l}{AUROC$\uparrow$} & \multicolumn{1}{l}{FPR95$\downarrow$} & \multicolumn{1}{l}{AUROC$\uparrow$} & \multicolumn{1}{l}{FPR95$\downarrow$} & \multicolumn{1}{l}{AUROC$\uparrow$} & \multicolumn{1}{l}{FPR95$\downarrow$} & \multicolumn{1}{l}{AUROC$\uparrow$} & \multicolumn{1}{l}{FPR95$\downarrow$} & \multicolumn{1}{l}{AUROC$\uparrow$} \\ \hline
                                                      & \multicolumn{12}{c}{$\pi=0.05$}                                                                                                                                                                                                                                                                                                                                                                                                                                       \\
OE                                                    & 28.44                               & 93.91                               & 24.74                               & 93.48                               & 8.99                                & 97.94                               & 29.44                               & 91.95                               & 38.32                               & 87.35                               & 38.43                               & 88.52                               \\
TaylorOE                                        & \textbf{15.11}                      & \textbf{96.99}                      & \textbf{4.89}                       & \textbf{98.82}                      & \textbf{1.71}                       & \textbf{99.60}                      & \textbf{5.71}                       & \textbf{98.73}                      & \textbf{22.19}                      & \textbf{94.98}                      & \textbf{18.66}                      & \textbf{95.76}                      \\ \hline
                                                      & \multicolumn{12}{c}{$\pi=0.1$}                                                                                                                                                                                                                                                                                                                                                                                                                                        \\
OE                                                    & 13.08                               & 97.00                               & 13.86                               & 96.84                               & 5.35                                & 98.79                               & 15.92                               & 96.34                               & 23.68                               & 92.53                               & 28.65                               & 91.94                               \\
TaylorOE                                        & \textbf{6.75}                       & \textbf{98.51}                      & \textbf{4.20}                       & \textbf{99.00}                      & \textbf{1.71}                       & \textbf{99.61}                      & \textbf{4.51}                       & \textbf{98.96}                      & \textbf{16.63}                      & \textbf{96.51}                      & \textbf{18.20}                      & \textbf{95.99}                      \\ \hline
                                                      & \multicolumn{12}{c}{$\pi=0.5$}                                                                                                                                                                                                                                                                                                                                                                                                                                        \\
OE                                                    & 2.50                                & 99.26                               & 3.78                                & 99.08                               & 2.18                                & 99.50                               & 4.09                                & 99.04                               & 10.41                               & 97.40                               & 16.81                               & 95.59                               \\
TaylorOE                                        & \textbf{2.11}                       & \textbf{99.26}                      & \textbf{2.55}                       & \textbf{99.25}                      & \textbf{1.78}                       & \textbf{99.57}                      & \textbf{2.52}                       & \textbf{99.26}                      & \textbf{8.07}                       & \textbf{98.31}                      & \textbf{14.27}                      & \textbf{96.69}                      \\ \hline
                                                      & \multicolumn{12}{c}{$\pi=0.75$}                                                                                                                                                                                                                                                                                                                                                                                                                                       \\
OE                                                    & 1.76    & \textbf{99.37}   & 2.47   & \textbf{99.28}   & \textbf{1.76}    & \textbf{99.55}    & 2.62     & 99.28     & 8.05     & 98.12          & 14.60                               & 96.32                               \\
TaylorOE                                        &  1.76   & 99.35    & \textbf{2.44}  &  99.27    &   1.87      &    99.54        &      \textbf{2.39}    &   \textbf{99.29}    & \textbf{7.31} & \textbf{98.51} & \textbf{13.69}  & \textbf{96.84}              \\ \hline
                                                      & \multicolumn{12}{c}{$\pi=0.9$}                                                                                                                                                                                                                                                                                                                                                                                                                                        \\
OE                                                    &  1.47                              & \textbf{99.46}                                &  \textbf{2.18}                               &  \textbf{99.32}                              &  \textbf{1.65}                               &  99.54                              & 2.09                                & 99.34                               &  6.90                               & 98.46                               &  13.11                              &  96.72                              \\
TaylorOE                                        & \textbf{1.39} & 99.41 & 2.20 & 99.31 & 1.78 & \textbf{99.55} & \textbf{2.01} & 99.34 & \textbf{6.64} & \textbf{98.65} & \textbf{12.77} &  \textbf{96.96}                                 \\ \hline
                                                      & \multicolumn{12}{c}{$\pi=1.0$}                                                                                                                                                                                                                                                                                                                                                                                                                                        \\
OE                                                    &  1.40                               & \textbf{99.45}                       &  \textbf{2.17}                               & \textbf{99.32}                     & \textbf{1.61}                        &  \textbf{99.56}                     &  2.06                                & 99.33                             &  6.23                                &  98.68                                & 12.40                            &  \textbf{97.03}                                   \\
TaylorOE                                        & \textbf{1.32} & 99.39 & 2.18 & 99.30 & 1.69 & 99.55 & \textbf{1.99} & 99.33 & \textbf{6.20} & \textbf{98.73} & \textbf{12.33} & 97.02       \\ \hline                         
\end{tabular}}
\end{table*}

%% file: tables/main_results_cifar100.tex
\begin{table*}[t]
\centering
\caption{Comparison between OE and TaylorOE with \texttt{CIFAR-100} as ID data.}
\label{tb:main_results_cifar100}
\scalebox{0.7}{
\begin{tabular}{l|cccccccccccc}
\hline
\multicolumn{1}{c|}{\multirow{3}{*}{\textbf{Method}}} & \multicolumn{12}{c}{\textbf{OOD Dataset}}                                                                                                                                                                                                                                                                                                                                                                                                                             \\
\multicolumn{1}{c|}{}                                 & \multicolumn{2}{c}{\textbf{SVHN}}                                         & \multicolumn{2}{c}{\textbf{LSUN-R}}                                       & \multicolumn{2}{c}{\textbf{LSUN-C}}                                       & \multicolumn{2}{c}{\textbf{iSUN}}                                         & \multicolumn{2}{c}{\textbf{Texture}}                                      & \multicolumn{2}{c}{\textbf{Places365}}                                    \\
\multicolumn{1}{c|}{}                                 & \multicolumn{1}{l}{FPR95$\downarrow$} & \multicolumn{1}{l}{AUROC$\uparrow$} & \multicolumn{1}{l}{FPR95$\downarrow$} & \multicolumn{1}{l}{AUROC$\uparrow$} & \multicolumn{1}{l}{FPR95$\downarrow$} & \multicolumn{1}{l}{AUROC$\uparrow$} & \multicolumn{1}{l}{FPR95$\downarrow$} & \multicolumn{1}{l}{AUROC$\uparrow$} & \multicolumn{1}{l}{FPR95$\downarrow$} & \multicolumn{1}{l}{AUROC$\uparrow$} & \multicolumn{1}{l}{FPR95$\downarrow$} & \multicolumn{1}{l}{AUROC$\uparrow$} \\ \hline
                                                      & \multicolumn{12}{c}{$\pi=0.05$}                                                                                                                                                                                                                                                                                                                                                                                                                                       \\
OE                                                    & 93.41                               & 61.76                               & 86.31                               & 70.77                               & 54.45                                & 88.29                               & 86.58                               & 71.90                               & 86.73                               & 68.11                               & 86.21                               & 70.57                               \\
TaylorOE                                        & \textbf{83.60}                      & \textbf{78.02}                      & \textbf{60.65}                       & \textbf{85.90}                      & \textbf{21.28}                       & \textbf{95.97}                      & \textbf{63.99}                       & \textbf{85.52}                      & \textbf{67.35}                      & \textbf{81.97}                      & \textbf{70.87}                      & \textbf{80.02}                      \\ \hline
                                                      & \multicolumn{12}{c}{$\pi=0.1$}                                                                                                                                                                                                                                                                                                                                                                                                                                        \\
OE                                                    & 92.35                               & 64.36                               & 81.71                               & 74.11                               & 44.70                                & 90.63                               & 82.24                               & 74.95                               & 83.73                               & 70.87                               & 83.36                               & 72.79                               \\
TaylorOE                                        & \textbf{80.59}                       & \textbf{80.40}                      & \textbf{40.65}                       & \textbf{91.75}                      & \textbf{16.08}                       & \textbf{97.04}                      & \textbf{44.76}                       & \textbf{90.94}                      & \textbf{60.16}                      & \textbf{84.76}                      & \textbf{62.86}                      & \textbf{83.13}                      \\ \hline
                                                      & \multicolumn{12}{c}{$\pi=0.5$}                                                                                                                                                                                                                                                                                                                                                                                                                                        \\
OE                                                    & 80.14                                & 78.20                               & 56.49                                & 86.01                               & 20.04                                & 96.17                               & 59.45                                & 85.73                               & 66.14                               & 81.68                               & 64.45                               & 82.20                               \\
TaylorOE                                        & \textbf{65.85}                       & \textbf{87.78}                      & \textbf{30.57}                       & \textbf{94.43}                      & \textbf{14.07}                       & \textbf{97.45}                      & \textbf{35.61}                       & \textbf{93.51}                      & \textbf{55.96}                       & \textbf{87.23}                      & \textbf{54.96}                      & \textbf{86.66}                      \\ \hline
                                                      & \multicolumn{12}{c}{$\pi=0.75$}                                                                                                                                                                                                                                                                                                                                                                                                                                       \\
OE                                                    & 73.43                               & 83.70                                & 51.15                               & 88.63                                & 16.64                               & 96.96                               & 55.10                                & 88.01                               & 60.74                                & 85.06                               & 58.21                               & 85.63                               \\
TaylorOE                                        & \textbf{61.11}                       & \textbf{89.38}                      & \textbf{32.46}                       & \textbf{94.17}                      & \textbf{15.13}                       & \textbf{97.36}                      & \textbf{37.33}                       & \textbf{93.33}                      & \textbf{55.10}                       & \textbf{87.75}                      & \textbf{54.79}                      & \textbf{87.30}                      \\ \hline
                                                      & \multicolumn{12}{c}{$\pi=0.9$}                                                                                                                                                                                                                                                                                                                                                                                                                                        \\
OE                                                    &  69.49                              & 86.55                                &  49.83                               &  89.70                              &  15.72                               &  97.21                              & 54.07                                & 88.93                               &  58.50                               & 86.70                               &  55.63                              &  87.41                              \\
TaylorOE                                        &  \textbf{61.42}                      &  \textbf{89.21}                    &  \textbf{31.23}                        &  \textbf{94.30}                    &  \textbf{14.70}                       &   \textbf{97.39}                   &  \textbf{36.20}                       & \textbf{93.45}                     &  \textbf{54.66}                        &  \textbf{87.81}                   & \textbf{54.25}                        &  \textbf{87.23}                                   \\ \hline
                                                      & \multicolumn{12}{c}{$\pi=1.0$}                                                                                                                                                                                                                                                                                                                                                                                                                                        \\
OE                                                    &  66.06                               & 88.41                       &  51.29                              & 89.60                     & 15.74                        &  97.28                     &  55.42                                & 88.68                             &  57.73                                &  87.47                                & 54.63                            &  88.34                                   \\
TaylorOE                                        &  \textbf{60.81}                      & \textbf{89.21}                               &  \textbf{30.27}                      &  \textbf{94.40}                              & \textbf{14.20}                                &  \textbf{97.43}                             &  \textbf{35.25}                      & \textbf{93.54}                      &  \textbf{54.39}                      &  \textbf{87.86}                     & \textbf{53.94}                      &  \textbf{87.19}      \\ \hline                            
\end{tabular}}
\end{table*}

%% file: tables/various_results2.tex
\begin{table*} \centering
    \caption{Performance comparison of TaylorOE and other OOD detection methods.}
    \label{tb:various_results}
    \resizebox{\textwidth}{!}{
    \Huge
        \begin{tabular}{c|c|*{2}{c}|*{2}{c}|*{2}{c}|*{2}{c}|*{2}{c}|*{2}{c}}
        \toprule
        \multirow{2}{*}{$\mathcal{D}_{\text{in}}$}&\multirow{2}{*}{Method}
        & \multicolumn{2}{c}{$\pi=0.05$} 
        & \multicolumn{2}{c}{$\pi=0.1$} 
        & \multicolumn{2}{c}{$\pi=0.5$} 
        & \multicolumn{2}{c}{$\pi=0.75$}  
        & \multicolumn{2}{c}{$\pi=0.9$}
        & \multicolumn{2}{c}{$\pi=1.0$}\\
                               &&  FPR95$\downarrow$ & AUROC$\uparrow$ 
                               &  FPR95$\downarrow$ & AUROC$\uparrow$ 
                               &  FPR95$\downarrow$ & AUROC$\uparrow$ 
                               &  FPR95$\downarrow$ & AUROC$\uparrow$ 
                               &  FPR95$\downarrow$ & AUROC$\uparrow$
                               &  FPR95$\downarrow$ & AUROC$\uparrow$\\
        \midrule
        \multirow{4}{*}{\textbf{CIFAR-10}} &Energy w/o $\mathcal{D}_\text{out}^\text{noisy}$ & 33.35 & 91.78 & 33.35 & 91.78 & 33.35 & 91.78 & 33.35 & 91.78 & 33.35 & 91.78 & 33.35 & 91.78 \\
        & OE                                    & 28.06 & 92.19 & 16.76 & 95.57 & 6.63 & 98.31  & 5.21 & 98.66  & 4.57 & 98.81  & 4.31 & \textbf{98.90}  \\
        & WOODS                                 & 17.77 & 96.27 & 11.89 & 97.53 & 5.80 & 98.56  & \textbf{4.80} & 98.66  & 4.56 & 98.68  & 4.41 & 98.68  \\
        & \cellcolor{red!25}TaylorOE   & \cellcolor{red!25}\textbf{11.38} & \cellcolor{red!25}\textbf{97.48} & \cellcolor{red!25}\textbf{8.67} & \cellcolor{red!25}\textbf{98.10}  & \cellcolor{red!25}\textbf{5.22} & \cellcolor{red!25}\textbf{98.72}  & \cellcolor{red!25}4.91 & \cellcolor{red!25}\textbf{98.80}  & \cellcolor{red!25}\textbf{4.46} & \cellcolor{red!25}\textbf{98.87}  & \cellcolor{red!25}\textbf{4.28} & \cellcolor{red!25}98.89  \\ 
        \midrule
        \multirow{4}{*}{\textbf{CIFAR-100}} &Energy w/o $\mathcal{D}_\text{out}^\text{noisy}$ & 74.67 & 79.49 & 74.67 & 79.49 & 74.67 & 79.49 & 74.67 & 79.49 & 74.67 & 79.49 & 74.67 & 79.49 \\
        & OE                                    & 82.28 & 71.90 & 78.01 & 74.62 & 57.78 & 85.00 & 52.54 & 88.00 & 50.54 & 89.41 & 50.14 & 89.96 \\
        & WOODS                                 & 65.92 & 83.48 & 59.87 & 85.99 & 47.96 & 89.98 & 46.25 & 90.49 & 45.45 & 90.72 & 47.75 & 89.84 \\
        & \cellcolor{red!25}TaylorOE   & \cellcolor{red!25}\textbf{61.29} & \cellcolor{red!25}\textbf{84.57} & \cellcolor{red!25}\textbf{50.85} & \cellcolor{red!25}\textbf{88.00} & \cellcolor{red!25}\textbf{42.83} & \cellcolor{red!25}\textbf{91.18} & \cellcolor{red!25}\textbf{42.65} & \cellcolor{red!25}\textbf{91.55} & \cellcolor{red!25}\textbf{42.08} & \cellcolor{red!25}\textbf{91.57} & \cellcolor{red!25}\textbf{41.48} & \cellcolor{red!25}\textbf{91.60}\\
        \bottomrule
    \end{tabular}
    }
\end{table*}

%% file: tables/extended_oe_results2.tex
\begin{table}[t]\centering
    \caption{Comparison between TaylorOE and OE with noisy auxiliary OOD data for Resampling~\cite{resampling}, DivOE~\cite{divoe}, and Normal setting.}
    \label{tb:extended_oe_results}
    \resizebox{0.48\textwidth}{!}{
    \large
    \begin{tabular}{c|c|*{2}{c}|*{2}{c}|*{2}{c}}
        \toprule
        \multirow{2}{*}{$\mathcal{D}_{\text{in}}$}&\multirow{2}{*}{Method} & \multicolumn{2}{c}{Normal} & \multicolumn{2}{c}{Resampling} &\multicolumn{2}{c}{DivOE} \\
        &  &  FPR95$\downarrow$ & AUROC$\uparrow$ &FPR95$\downarrow$ & AUROC$\uparrow$&FPR95$\downarrow$ & AUROC$\uparrow$\\
        \midrule
        \multirow{2}{*}{\textbf{CIFAR-10}} & OE & 8.97 & 97.85 & 8.75 & 97.88 & 9.18 & 97.83 \\
        & TaylorOE &\textbf{6.61} &\textbf{98.45} & \textbf{6.54} &\textbf{98.47} & \textbf{6.37} &\textbf{98.47}\\
        \midrule
        \multirow{2}{*}{\textbf{CIFAR-100}} & OE & 54.56 & 85.91 & 54.97 & 85.87 & 56.90 & 85.62 \\
        & TaylorOE &\textbf{45.48} &\textbf{90.89} & \textbf{44.66} &\textbf{91.04} & \textbf{47.64} &\textbf{90.11}\\
        \bottomrule
    \end{tabular}
    }
\end{table}

%% file: sec/6_discussion.tex
\section{Discussion}
\label{sec:discussion}
\begin{figure}[t]
    \centering
    \begin{minipage}{0.49\hsize}
        \centering
        \includegraphics[width=1.0\linewidth]{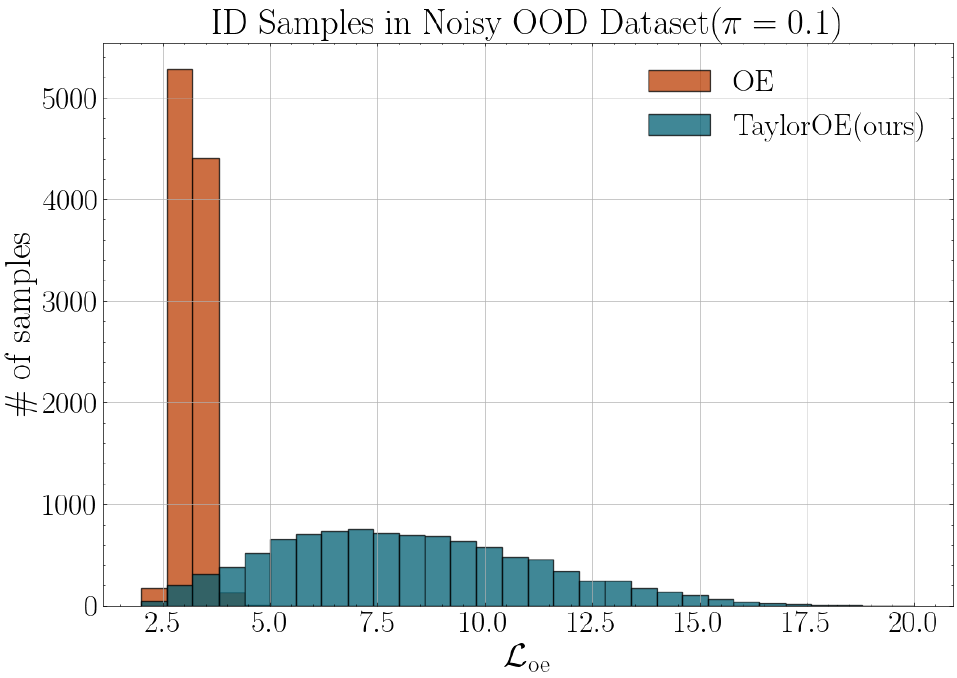}
    \end{minipage}
    \begin{minipage}{0.49\hsize}
        \centering
        \includegraphics[width=1.0\linewidth]{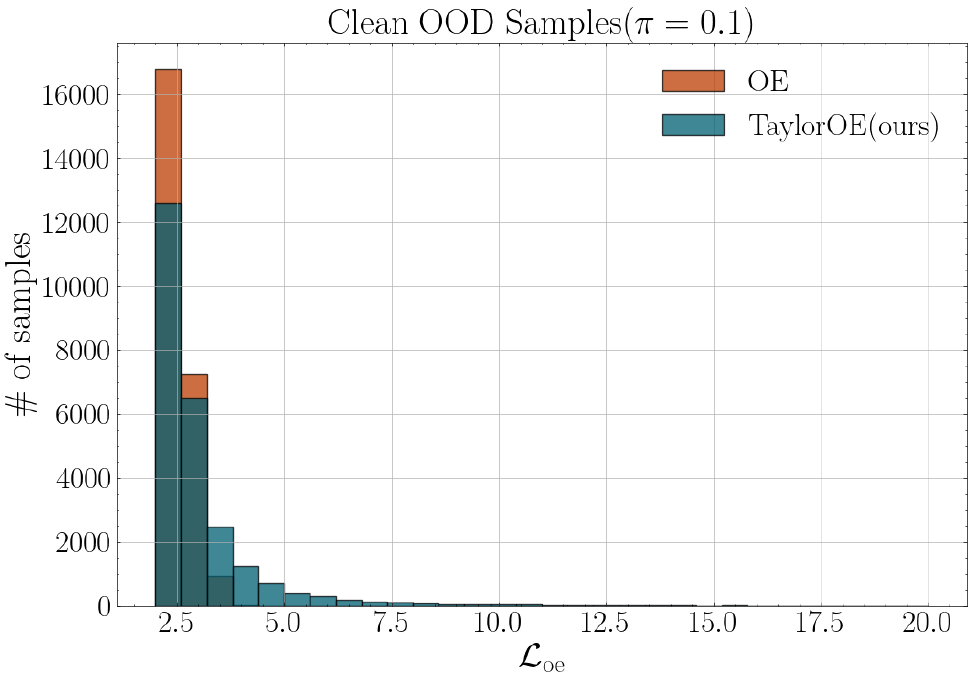}
    \end{minipage}
    \caption{Histograms of $\mathcal{L}_{\text{oe}}$~(Eq.~\ref{l_oe}) for ID samples in the noisy OOD dataset(left) and clean OOD samples(right). When training via OE, both $\mathcal{L}_{\text{oe}}$ values for ID data in the noisy OOD dataset and clean OOD data show low values, indicating that both of these data affect the OE training. In contrast, TaylorOE maintains higher $\mathcal{L}_{\text{oe}}$ values for ID data in the noisy OOD dataset while showing low values for clean OOD data, indicating that TaylorOE can train clean OOD data while mitigating the influence of ID data in the noisy OOD dataset on the training process.}
    \label{fig:loe_hist}
\end{figure}

\begin{figure}[t]
    \centering
    \begin{minipage}{0.49\hsize}
        \centering
        \includegraphics[width=1.0\linewidth]{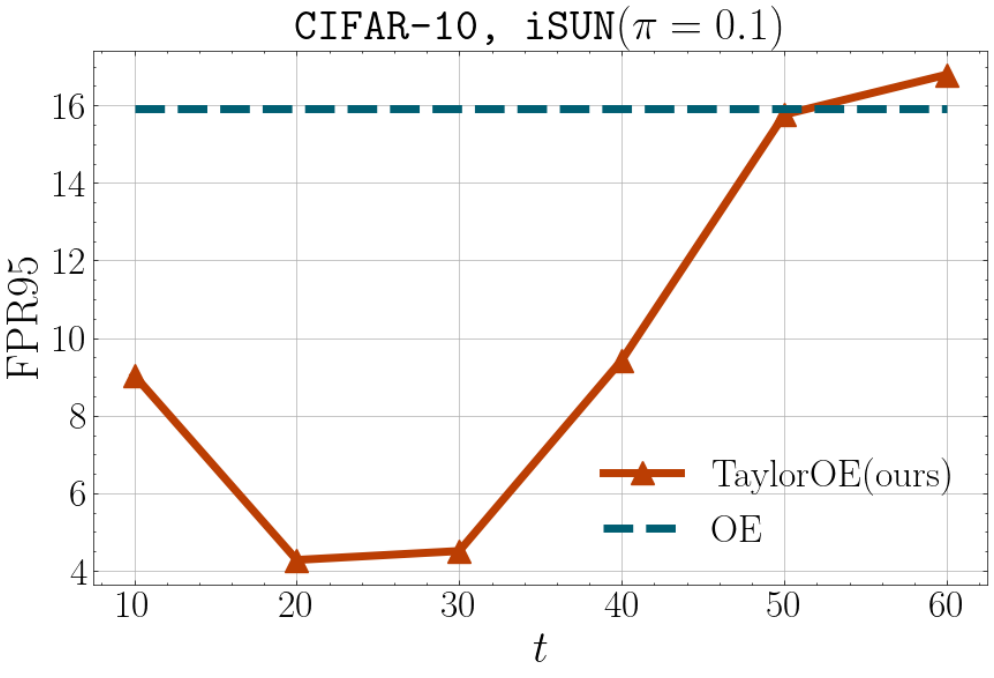}
    \end{minipage}
    \begin{minipage}{0.49\hsize}
        \centering
        \includegraphics[width=1.0\linewidth]{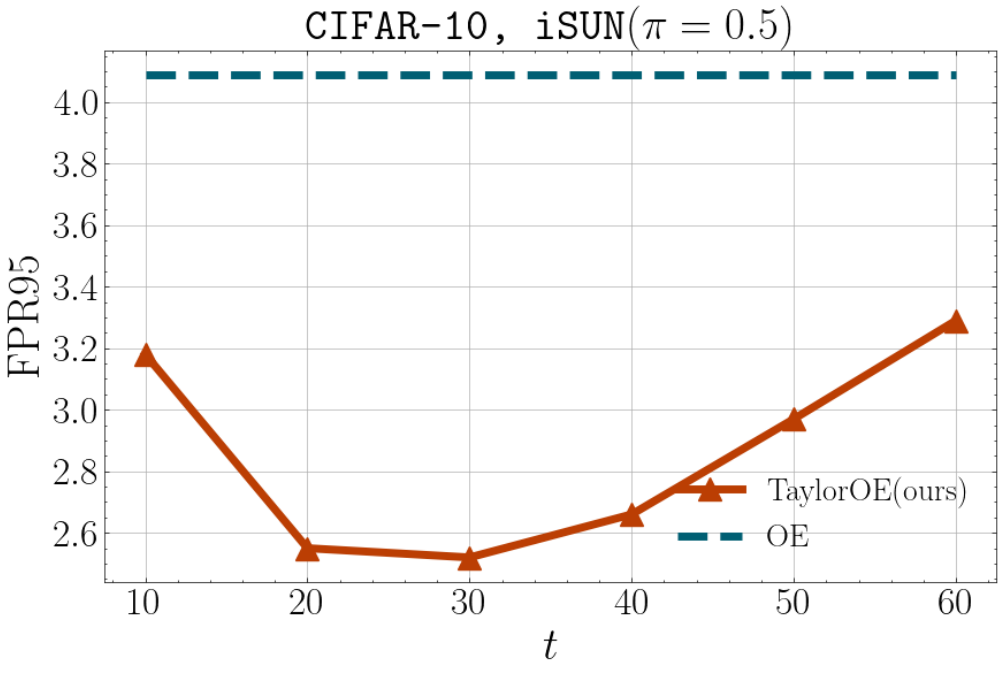}
    \end{minipage}
    \caption{Relationship between noise ratio $\pi$ and TaylorOE parameter $t$ for \texttt{CIFAR-10}.}
    \label{fig:discussion_cifar10}
\end{figure}
\begin{figure}[t]
    \centering
    \begin{minipage}{0.49\hsize}
        \centering
        \includegraphics[width=1.0\linewidth]{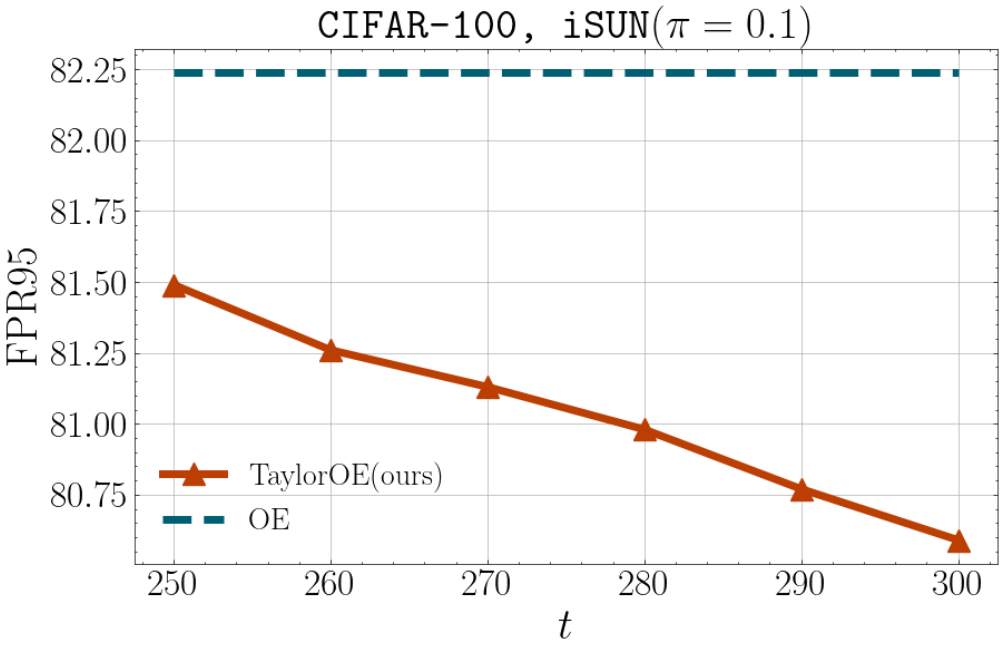}
    \end{minipage}
    \begin{minipage}{0.49\hsize}
        \centering
        \includegraphics[width=1.0\linewidth]{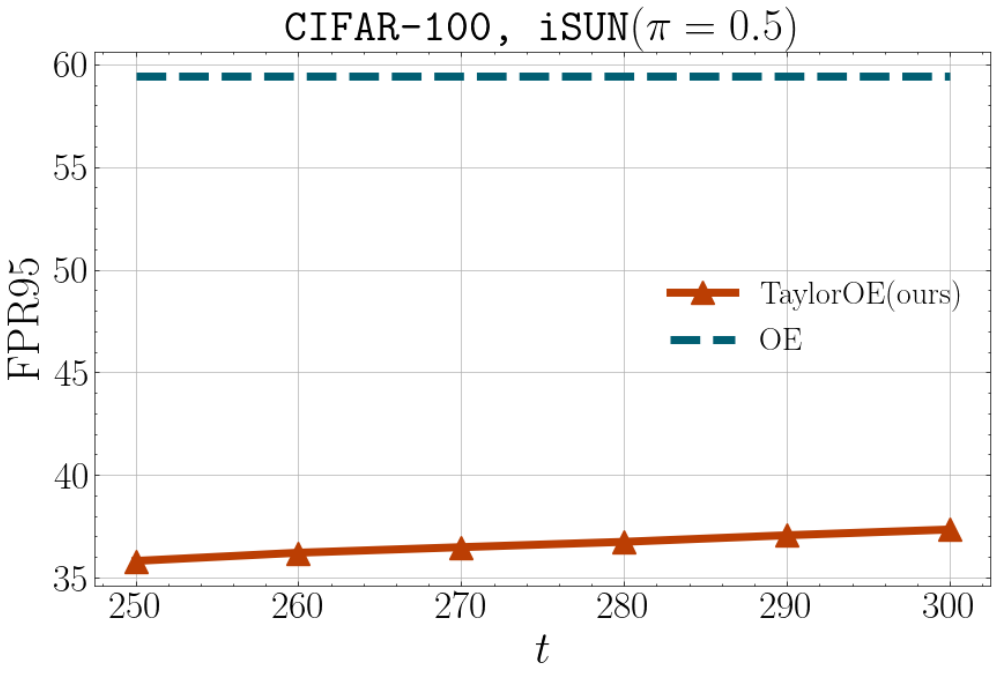}
    \end{minipage}
    \caption{Relationship between noise ratio $\pi$ and TaylorOE parameter $t$ for \texttt{CIFAR-100}.}
    \label{fig:discussion_cifar100}
\end{figure}

\noindent\textbf{Impact on OE's Training:}
Fig.~\ref{fig:loe_hist} shows histograms of $\mathcal{L}_{\text{oe}}$~(Eq.~\ref{l_oe}) for ID data in the noisy auxiliary OOD dataset, namely \texttt{CIFAR-10}, and clean OOD data, which consists of 25,000 randomly selected samples from \texttt{300K Random Images}. These values were obtained from fine-tuned models with the noisy OOD dataset~($\pi=0.1$) using both OE and TaylorOE. As shown in Fig.~\ref{fig:logit_graph}, when data influence model training, $\mathcal{L}_{\text{oe}}$ values decrease. Under OE training, both ID data and clean OOD data in the noisy auxiliary OOD dataset show low $\mathcal{L}_{\text{oe}}$ values, suggesting a pronounced impact of ID data in the noisy auxiliary OOD dataset on model training. In contrast, TaylorOE maintains higher $\mathcal{L}_{\text{oe}}$ values for ID data in the noisy auxiliary OOD dataset while showing low values for clean OOD data, indicating that TaylorOE effectively balances training from clean OOD data while reducing the impact of ID data in the noisy auxiliary OOD dataset on the training process. From these results, we found that, compared to the original OE, TaylorOE facilitates more robust OE training by mitigating the detrimental influence of ID data in the noisy auxiliary OOD dataset.

\noindent\textbf{Noise ratio $\pi$:}
We analyze the relationship between the proportion of ID data in the noisy auxiliary OOD dataset and the hyperparameter $t$ of the proposed method. Fig.~\ref{fig:discussion_cifar10} compares the results obtained when training with the noisy OOD dataset at noise ratios of $\pi=0.1$ and $\pi=0.5$ using \texttt{CIFAR-10} as ID data and \texttt{iSUN} as test OOD data. We narrowed the $t$ intervals to $\{10,20,30,40,50\}$ to examine accuracy variations. When $\pi=0.1$~(90\% ID data), the lowest FPR95 value occurs at $t=20$; beyond this value, FPR95 increases. For $\pi=0.5$~(50\% ID data), the boundary shifts to $t=30$, indicating that as $\pi$ increases and the amount of ID data in the noisy auxiliary OOD dataset decreases, the negative impact of ID data on training diminishes. Consequently, there is less need for robust regularization, and a higher $t$ value yields better detection accuracy.

\noindent\textbf{ID Dataset $\mathcal{D}_{\text{in}}$:}
We discuss the relationship between ID data and the hyperparameter $t$ of the proposed method. Fig.~\ref{fig:discussion_cifar100} uses \texttt{CIFAR-100} as ID data. For \texttt{CIFAR-10}, the optimal value of $t$ is 20 when $\pi=0.1$ and 30 when $\pi=0.5$, indicating a significant relationship between $\pi$ and $t$. However, the optimal $t$ value also depends on the number of classes of ID data and the properties of the test OOD data. When \texttt{CIFAR-100} was used as the ID data, the larger number of classes resulted in lower overall probability values, meaning that setting $t$ too low can reduce the effect on clean OOD data. Therefore, we explored values of $t$ in the range $\{250,260,270,280,290,300\}$. We found that $t=300$ yields the lowest FPR95 value for both $\pi=0.1$ and $\pi=0.5$. Additionally, increasing $t$ further may lead to an even lower FPR95 value. In our experiments, we optimized $t$ based on the OOD detection performance of the validation set; however, as the peak detection performance varies by dataset, hyperparameter tuning with consideration of this variability is necessary when applying our method.

%% file: sec/7_conclusion.tex
\section{Conclusion}
\label{sec:conclusion}
This paper proposed TaylorOE, a regularization method for handling noisy auxiliary OOD data that enables the direct use of mixed ID and OOD data without additional sampling or cleaning. Through various experiments with noisy auxiliary OOD datasets, we demonstrated the limitations of OE, the effectiveness of the proposed regularization, and the applicability of TaylorOE to advanced OE methods, such as OOD sampling and synthesis. Our polynomial regularization term derived from Taylor expansion could contribute to a better understanding of loss functions in OOD detection and the enhanced utilization of OOD data for generalization. However, our analysis reveals sensitivity in the Taylor expansion dimension to noise ratio and the number of ID classes, which complicates hyperparameter tuning. These limitations will be addressed in future works.

%% file: main.bbl
\begin{thebibliography}{37}
\providecommand{\natexlab}[1]{#1}
\providecommand{\url}[1]{\texttt{#1}}
\expandafter\ifx\csname urlstyle\endcsname\relax
  \providecommand{\doi}[1]{doi: #1}\else
  \providecommand{\doi}{doi: \begingroup \urlstyle{rm}\Url}\fi

\bibitem[Chen et~al.(2021)Chen, Li, Wu, Liang, and Jha]{atom}
Jiefeng Chen, Yixuan Li, Xi Wu, Yingyu Liang, and Somesh Jha.
\newblock Atom: Robustifying out-of-distribution detection using outlier mining.
\newblock In \emph{Machine Learning and Knowledge Discovery in Databases. Research Track: European Conference, ECML PKDD 2021, Bilbao, Spain, September 13--17, 2021, Proceedings, Part III 21}, pages 430--445. Springer, 2021.

\bibitem[Cimpoi et~al.(2014)Cimpoi, Maji, Kokkinos, Mohamed, and Vedaldi]{texture}
Mircea Cimpoi, Subhransu Maji, Iasonas Kokkinos, Sammy Mohamed, and Andrea Vedaldi.
\newblock Describing textures in the wild.
\newblock In \emph{CVPR}, pages 3606--3613, 2014.

\bibitem[DeVries and Taylor(2018)]{conf_learning}
Terrance DeVries and Graham~W Taylor.
\newblock Learning confidence for out-of-distribution detection in neural networks.
\newblock \emph{arXiv preprint arXiv:1802.04865}, 2018.

\bibitem[Djurisic et~al.(2022)Djurisic, Bozanic, Ashok, and Liu]{ash}
Andrija Djurisic, Nebojsa Bozanic, Arjun Ashok, and Rosanne Liu.
\newblock Extremely simple activation shaping for out-of-distribution detection.
\newblock \emph{arXiv preprint arXiv:2209.09858}, 2022.

\bibitem[Du et~al.(2022)Du, Wang, Cai, and Li]{vos}
Xuefeng Du, Zhaoning Wang, Mu Cai, and Yixuan Li.
\newblock Vos: Learning what you don't know by virtual outlier synthesis.
\newblock In \emph{ICLR}, 2022.

\bibitem[Du et~al.(2023)Du, Sun, Zhu, and Li]{dream}
Xuefeng Du, Yiyou Sun, Jerry Zhu, and Yixuan Li.
\newblock Dream the impossible: Outlier imagination with diffusion models.
\newblock In \emph{NIPS}, pages 60878--60901. Curran Associates, Inc., 2023.

\bibitem[Feng et~al.(2021)Feng, Shu, Lin, Lv, Li, and An]{taylor_ce}
Lei Feng, Senlin Shu, Zhuoyi Lin, Fengmao Lv, Li Li, and Bo An.
\newblock Can cross entropy loss be robust to label noise?
\newblock In \emph{IJCAI}, pages 2206--2212, 2021.

\bibitem[Hendrycks and Gimpel(2017)]{baseline}
Dan Hendrycks and Kevin Gimpel.
\newblock A baseline for detecting misclassified and out-of-distribution examples in neural networks.
\newblock In \emph{ICLR}, 2017.

\bibitem[Hendrycks et~al.(2019)Hendrycks, Mazeika, and Dietterich]{outlier_exposure}
Dan Hendrycks, Mantas Mazeika, and Thomas Dietterich.
\newblock Deep anomaly detection with outlier exposure.
\newblock In \emph{ICLR}, 2019.

\bibitem[Hendrycks et~al.(2022)Hendrycks, Basart, Mazeika, Zou, Kwon, Mostajabi, Steinhardt, and Song]{maxlogit}
Dan Hendrycks, Steven Basart, Mantas Mazeika, Andy Zou, Joseph Kwon, Mohammadreza Mostajabi, Jacob Steinhardt, and Dawn Song.
\newblock Scaling out-of-distribution detection for real-world settings.
\newblock In \emph{ICLR}, pages 8759--8773. PMLR, 2022.

\bibitem[Huang et~al.(2021)Huang, Geng, and Li]{gradnorm}
Rui Huang, Andrew Geng, and Yixuan Li.
\newblock On the importance of gradients for detecting distributional shifts in the wild.
\newblock In \emph{NIPS}, pages 677--689. Curran Associates, Inc., 2021.

\bibitem[Huber(1992)]{huber}
Peter~J Huber.
\newblock Robust estimation of a location parameter.
\newblock In \emph{Breakthroughs in statistics: Methodology and distribution}, pages 492--518. Springer, 1992.

\bibitem[Jiang et~al.(2024)Jiang, Cheng, Chen, Wang, and Wei]{dos}
Wenyu Jiang, Hao Cheng, MingCai Chen, Chongjun Wang, and Hongxin Wei.
\newblock {DOS}: Diverse outlier sampling for out-of-distribution detection.
\newblock In \emph{ICLR}, 2024.

\bibitem[Katz-Samuels et~al.(2022)Katz-Samuels, Nakhleh, Nowak, and Li]{woods}
Julian Katz-Samuels, Julia~B Nakhleh, Robert Nowak, and Yixuan Li.
\newblock Training ood detectors in their natural habitats.
\newblock In \emph{ICML}, pages 10848--10865. PMLR, 2022.

\bibitem[Krizhevsky and Hinton(2009)]{cifar}
Alex Krizhevsky and Geoffrey Hinton.
\newblock Learning multiple layers of features from tiny images.
\newblock Technical Report~0, University of Toronto, Toronto, Ontario, 2009.

\bibitem[Lee et~al.(2018)Lee, Lee, Lee, and Shin]{mahalanobis}
Kimin Lee, Kibok Lee, Honglak Lee, and Jinwoo Shin.
\newblock A simple unified framework for detecting out-of-distribution samples and adversarial attacks.
\newblock In \emph{NIPS}. Curran Associates, Inc., 2018.

\bibitem[Leng et~al.(2022)Leng, Tan, Liu, Cubuk, Shi, Cheng, and Anguelov]{polyloss}
Zhaoqi Leng, Mingxing Tan, Chenxi Liu, Ekin~Dogus Cubuk, Jay Shi, Shuyang Cheng, and Dragomir Anguelov.
\newblock Polyloss: A polynomial expansion perspective of classification loss functions.
\newblock In \emph{International Conference on Learning Representations}, 2022.

\bibitem[Li and Vasconcelos(2020)]{resampling}
Yi Li and Nuno Vasconcelos.
\newblock Background data resampling for outlier-aware classification.
\newblock In \emph{CVPR}, pages 13215--13224. Computer Vision Foundation / {IEEE}, 2020.

\bibitem[Liang et~al.(2018)Liang, Li, and Srikant]{odin}
Shiyu Liang, Yixuan Li, and R. Srikant.
\newblock Enhancing the reliability of out-of-distribution image detection in neural networks.
\newblock In \emph{ICLR}, 2018.

\bibitem[Lin et~al.(2021)Lin, Roy, and Li]{mood}
Ziqian Lin, Sreya~Dutta Roy, and Yixuan Li.
\newblock Mood: Multi-level out-of-distribution detection.
\newblock In \emph{CVPR}, pages 15313--15323, 2021.

\bibitem[Liu et~al.(2020)Liu, Wang, Owens, and Li]{energy}
Weitang Liu, Xiaoyun Wang, John Owens, and Yixuan Li.
\newblock Energy-based out-of-distribution detection.
\newblock In \emph{NIPS}, pages 21464--21475. Curran Associates, Inc., 2020.

\bibitem[Liu and Guo(2020)]{peer_loss}
Yang Liu and Hongyi Guo.
\newblock Peer loss functions: Learning from noisy labels without knowing noise rates.
\newblock In \emph{ICML}, pages 6226--6236. PMLR, 2020.

\bibitem[Loshchilov and Hutter(2017)]{cosine_scheduler}
Ilya Loshchilov and Frank Hutter.
\newblock {SGDR}: Stochastic gradient descent with warm restarts.
\newblock In \emph{ICLR}, 2017.

\bibitem[Ming et~al.(2022)Ming, Fan, and Li]{poem}
Yifei Ming, Ying Fan, and Yixuan Li.
\newblock Poem: Out-of-distribution detection with posterior sampling.
\newblock In \emph{ICML}, pages 15650--15665. PMLR, 2022.

\bibitem[Netzer et~al.(2011)Netzer, Wang, Coates, Bissacco, Wu, Ng, et~al.]{svhn}
Yuval Netzer, Tao Wang, Adam Coates, Alessandro Bissacco, Baolin Wu, Andrew~Y Ng, et~al.
\newblock Reading digits in natural images with unsupervised feature learning.
\newblock In \emph{NIPS workshop on deep learning and unsupervised feature learning}, page~7. Granada, Spain, 2011.

\bibitem[Sharifi et~al.(2025)Sharifi, Entesari, Safaei, Patel, and Fazlyab]{grad_reg}
Sina Sharifi, Taha Entesari, Bardia Safaei, Vishal~M Patel, and Mahyar Fazlyab.
\newblock Gradient-regularized out-of-distribution detection.
\newblock In \emph{ECCV}, pages 459--478. Springer, 2025.

\bibitem[Sun and Li(2022)]{dice}
Yiyou Sun and Yixuan Li.
\newblock Dice: Leveraging sparsification for out-of-distribution detection.
\newblock In \emph{ECCV}, pages 691--708. Springer, 2022.

\bibitem[Tao et~al.(2023)Tao, Du, Zhu, and Li]{npos}
Leitian Tao, Xuefeng Du, Jerry Zhu, and Yixuan Li.
\newblock Non-parametric outlier synthesis.
\newblock In \emph{ICLR}, 2023.

\bibitem[Wang et~al.(2023)Wang, Ye, Liu, Dai, Kalander, Liu, HAO, and Han]{doe}
Qizhou Wang, Junjie Ye, Feng Liu, Quanyu Dai, Marcus Kalander, Tongliang Liu, Jianye HAO, and Bo Han.
\newblock Out-of-distribution detection with implicit outlier transformation.
\newblock In \emph{ICLR}, 2023.

\bibitem[Wei et~al.(2022)Wei, Xie, Cheng, Feng, An, and Li]{logitnorm}
Hongxin Wei, Renchunzi Xie, Hao Cheng, Lei Feng, Bo An, and Yixuan Li.
\newblock Mitigating neural network overconfidence with logit normalization.
\newblock In \emph{ICML}, pages 23631--23644. PMLR, 2022.

\bibitem[Xu et~al.(2015)Xu, Ehinger, Zhang, Finkelstein, Kulkarni, and Xiao]{isun}
Pingmei Xu, Krista~A Ehinger, Yinda Zhang, Adam Finkelstein, Sanjeev~R Kulkarni, and Jianxiong Xiao.
\newblock Turkergaze: Crowdsourcing saliency with webcam based eye tracking.
\newblock \emph{arXiv preprint arXiv:1504.06755}, 2015.

\bibitem[Yu et~al.(2015)Yu, Seff, Zhang, Song, Funkhouser, and Xiao]{lsun}
Fisher Yu, Ari Seff, Yinda Zhang, Shuran Song, Thomas Funkhouser, and Jianxiong Xiao.
\newblock Lsun: Construction of a large-scale image dataset using deep learning with humans in the loop.
\newblock \emph{arXiv preprint arXiv:1506.03365}, 2015.

\bibitem[Zagoruyko and Komodakis(2016)]{wideresnet}
Sergey Zagoruyko and Nikos Komodakis.
\newblock Wide residual networks.
\newblock In \emph{BMVC}. {BMVA} Press, 2016.

\bibitem[Zhang et~al.(2023)Zhang, Inkawhich, Linderman, Chen, and Li]{MixOE}
Jingyang Zhang, Nathan Inkawhich, Randolph Linderman, Yiran Chen, and Hai Li.
\newblock Mixture outlier exposure: Towards out-of-distribution detection in fine-grained environments.
\newblock In \emph{WACV}, pages 5531--5540, 2023.

\bibitem[Zhang and Sabuncu(2018)]{generalized_ce}
Zhilu Zhang and Mert Sabuncu.
\newblock Generalized cross entropy loss for training deep neural networks with noisy labels.
\newblock In \emph{NIPS}. Curran Associates, Inc., 2018.

\bibitem[Zhou et~al.(2017)Zhou, Lapedriza, Khosla, Oliva, and Torralba]{places365}
Bolei Zhou, Agata Lapedriza, Aditya Khosla, Aude Oliva, and Antonio Torralba.
\newblock Places: A 10 million image database for scene recognition.
\newblock \emph{IEEE TPAMI}, 40\penalty0 (6):\penalty0 1452--1464, 2017.

\bibitem[Zhu et~al.(2023)Zhu, Geng, Yao, Liu, Niu, Sugiyama, and Han]{divoe}
Jianing Zhu, Yu Geng, Jiangchao Yao, Tongliang Liu, Gang Niu, Masashi Sugiyama, and Bo Han.
\newblock Diversified outlier exposure for out-of-distribution detection via informative extrapolation.
\newblock In \emph{NIPS}, pages 22702--22734. Curran Associates, Inc., 2023.

\end{thebibliography}
